\documentclass{article}

\usepackage[final]{neurips_2022_ml4ps}

\usepackage[utf8]{inputenc} 
\usepackage[T1]{fontenc}    
\usepackage{hyperref}       
\usepackage{url}            
\usepackage{booktabs}       
\usepackage{amsfonts}       
\usepackage{nicefrac}       
\usepackage{microtype}      
\usepackage{xcolor}         
\usepackage{bm}
\usepackage{graphicx}
\usepackage{mathtools}

\usepackage{xspace}

\makeatletter
\DeclareRobustCommand\onedot{\futurelet\@let@token\@onedot}
\def\@onedot{\ifx\@let@token.\else.\null\fi\xspace}

\def\eg{\emph{e.g}\onedot}

\def\etc{\emph{etc}\onedot}

\makeatother

\usepackage[linesnumbered,ruled,vlined]{algorithm2e}

\title{Differentiable Physics-based Greenhouse Simulation}

\author{%
  Nhat M. Nguyen \\
  Koidra.ai\\
  \texttt{nhat.nguyen@koidra.ai} \\
  \And
  Hieu T. Tran \\
  Koidra.ai\\
  \texttt{hieu.tran@koidra.ai} \\
  \And
  Minh V. Duong \\
  Koidra.ai\\
  \texttt{minh.duong@koidra.ai} \\
  \And
  Hanh Bui \\
  Koidra.ai\\
  \texttt{hanh.bui@koidra.ai} \\
  \And
  Kenneth Tran \\
  Koidra.ai \\
  \texttt{ken@koidra.ai} \\
}

\begin{document}

\maketitle

\begin{abstract}
  We present a differentiable greenhouse simulation model based on physical processes whose parameters can be obtained by training from real data.
  The physics-based simulation model is fully interpretable and is able to do state prediction for both climate and crop dynamics in the greenhouse over very a long time horizon. 
  The model works by constructing a system of linear differential equations and solving them to obtain the next state. 
  We propose a procedure to solve the differential equations, handle the problem of missing unobservable states in the data, and train the model efficiently. 
  Our experiment shows that the procedure is effective. 
  The model improves significantly after training and can simulate a greenhouse that grows cucumbers accurately.
\end{abstract}

\section{Introduction}
\label{intro}
There has been a growing interest in applying machine learning in various agricultural domains and such attempts have achieved multiple successes to allow for more efficient and better farming (\cite{meshram2021machine}, \cite{yang2021applications}). 
With the world population still growing, food security remains to be one of the most important issues for many countries and machine learning applications in agriculture and horticulture can help alleviate this issue. 
One of such applications is to use machine learning to improve optimal control in greenhouses.
For tuning and learning of the optimal controllers in the greenhouses, there is a need to use simulators, as it is prohibitively expensive to do so in the real systems.
The most prominent and successful class of models to simulate the greenhouses are based on physical processes.
Physics-based greenhouse models are well-studied (\cite{vanthoor2011model}, \cite{marcelis1994simulation}, \cite{wells1992modelling}) and have been used for optimal greenhouse control, for example, in \cite{chen2022intelligent}.
However, those models are usually fixed, and their parameters have to be hand-tuned for each greenhouse.
This limits the rapid deployment and adaptation of such models in commercial settings.
On the other hand, differentiable physics (\cite{ramsundar2021differentiable}, \cite{thuerey2021pbdl}) is an emerging interdisciplinary paradigm where automatic differentiation \cite{neidinger2010introduction} is applied in the numerical simulation of physical systems and has the potential to facilitate automatic learning of physical systems parameters.
Furthermore, many optimal control algorithms, for example, the Direct Multiple Shooting method \cite{diehl2006fast}, require not only an accurate model of the dynamics but also the Jacobian and Hessian of the dynamic with respect to the state vector.
Thus, there is a need to develop an accurate and differentiable physics-based greenhouse simulation implementation that can be trained end-to-end in order to address those challenges and allow better adoption of machine learning in greenhouse optimal control.
This work presents such a greenhouse model implementation and proposes an efficient training procedure to address various problems when training the model.
\par
The contributions of this paper are as follow: 
\textbf{(1)} We present a fully interpretable and differentiable physics-based greenhouse model implementation that can be trained end-to-end and can be used for accurate simulation of the greenhouse and for optimal greenhouse control. 
\textbf{(2)} We propose an efficient and numerically stable way to solve the linear system of differential equations for greenhouse dynamics, which can be used for similar simulation systems that use differential equations.
\textbf{(3)} We propose an efficient and effective training procedure to handle difficulties in training long sequential greenhouse data.
\textbf{(4)} As a proof of concept, we show some good preliminary results in predicting the cumulative weight and number of fruits harvested in a greenhouse that grows cucumbers.

\section{Physics-based Greenhouse Simulation Model}
\subsection{Greenhouse simulation model formulation}

\begin{figure}
  \centering
  \includegraphics[width=1.0\linewidth]{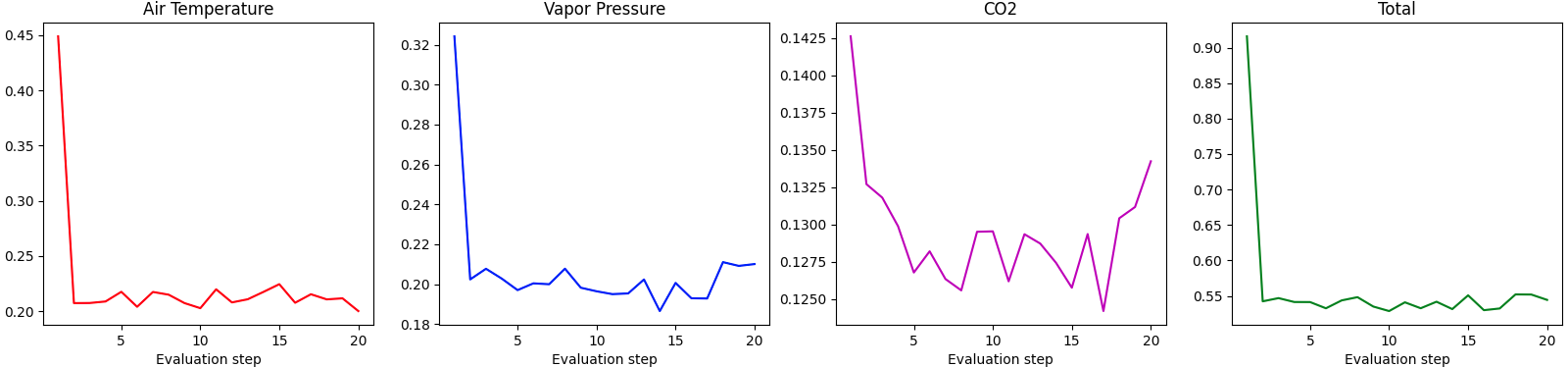}
  \caption{Individual loss components and total loss of the climate states on Croperator dataset.}
  \label{fig:climate_loss_plot}
\end{figure}

Let  
$\mathcal{D} = \{(\bm{x}_{i},\bm{u}_{i})|i=0..N\}$ 
be the sequential trajectory of a greenhouse with length $N$ in a single growing season.
$\bm{x}_i$ is the $D_{\bm{x}}$-dimension state vector and $\bm{u}_i$ is the $D_{\bm{u}}$-dimension control vector at time step $t$. 
The time steps are divided equally in time, with $\Delta t$ being the interval between time steps.

\par

The state vector $\bm{x}_i$ includes both climate states and crop states. 
Climate states are states that are related or contribute to the climate dynamic in the greenhouse.
For example: air temperature, air humidity, how much CO2 is in the air, \etc
Crop states are states that are related to the plants.
For example: average plant temperature in the last 24 hours, carbohydrate stored in the leaf, fruits, \etc
However, many states are unobservable, either due to the lack of sensors or due to the impossibility of measuring them.
For example, a greenhouse may not have sensors to measure the temperature of the plant canopy, and it is considered an unobservable state. 
Another example is that it is not possible to quantify the amount of carbohydrate in the plant stem without causing harm to the plant and this also an unobservable state.
The unobservable states are not available in the data collection process but may contribute significantly to the dynamics of a greenhouse.
To accurately simulate the dynamics of a greenhouse, it is important to handle the missing states.
A process for handling missing states during training is described in the experiment section \ref{subsection:training_details}.
The control vector $\bm{u}_i$ also includes control for both climate and crop. 
For example, setpoints of the heating pipes, setpoints of the fogging system, or harvesting action of the fruits.
\par

A greenhouse simulator $\mathcal{S}_{\theta}$, parameterized by the parameter vector $\theta$, calculates a prediction of $T$ next states ${\bm{x}}_{i+1},...,{\bm{x}}_{i+T}$ given the starting state $\bm{x}_{i}$ and the control vectors $\bm{u}_{i},...,\bm{u}_{i+T-1}$ using the recursive formulation:
$\bm{x}_{k+1}=\mathcal{S}_{\theta}(\bm{x}_{k},\bm{u}_{k}) \text{ for } k=i \rightarrow i+T-1$.
The physics-based greenhouse simulation model predicts the next state $\bm{x}_{k+1}$ by solving a system of linear differential equations:

\begin{equation}
\label{eq:ode_original}
    \dot{\bm{x}}=A_{\theta}(\bm{x},\bm{u}) \bm{x} + b_{\theta}(\bm{x},\bm{u})
\end{equation}

where $\dot{\bm{x}}$ is the derivative of the state vector $\bm{x}$ over time. $A_{\theta}(\bm{x},\bm{u})$ is a $D_{\bm{x}} \times D_{\bm{x}}$ matrix and $b_{\theta}(\bm{x},\bm{u})$ is a $D_{\bm{x}}$ dimensional vector.
Both terms are functions of the state vector $\bm{x}$ and the control vector $\bm{u}$, and are parameterized by $\theta$.
The parameter vector $\theta$ consists of parameters that affect the physical and biological processes in the greenhouse.
Such parameterization scheme is fully interpretable while also allows human experts to contribute to the process of tuning the model.
Examples of physical processes parameters include the height of the greenhouse, the normalized area of the ventilation, parameters of greenhouse screens, the capacity of the fogging system, \etc
The biological processes parameters depend on the type of crop being grown. 
For vine crops (\eg cucumber, tomato, \etc), examples are maximum leaf area index, light absorption coefficient of the canopy, \etc
\par
The next state $\bm{x}_{k+1}$ is obtained by substituting $\bm{x}_{k}$ into (\ref{eq:ode_original}) and solve it for time step $\Delta t$. 
Detail how to solve (\ref{eq:ode_original}) is outlined in \ref{subsection:solve_ode}.
The outstanding task of the simulator is to find $A_{\theta}(\bm{x},\bm{u})$ and $b_{\theta}(\bm{x},\bm{u})$. To construct them, we follow the physical and biological processes of vine crops described in \cite{vanthoor2011model}. 
The parameters in $\theta$ are also obtained from \cite{vanthoor2011model}.
To make our simulator differentiable, we implemented both the construction of the terms in (\ref{eq:ode_original}) as well as the solving of (\ref{eq:ode_original}) in Pytorch \cite{NEURIPS2019_9015}.
\par
The training of our simulator reduces to finding the best values for the parameters in $\theta$.
In practice, those parameters can only take values in physically feasible ranges.
Therefore, we further formulate the parameters in $\theta$ using the sigmoid function $\sigma$: $\theta_j = \theta^{min}_j + \sigma(\theta^*_j) (\theta^{max}_j - \theta^{min}_j)$.
This constrains $\theta_j \in (\theta^{min}_j, \theta^{max}_j)$. 
In this paper, we consider the greenhouses where cucumber is grown. 
$(\theta^{min}_j, \theta^{max}_j)$ are chosen to be from 0.5 to 2.0 times the value of each parameter in \cite{vanthoor2011model} for parameters of the climate processes and in \cite{marcelis1994simulation} for parameters of the cucumber crop model.

\subsection{Solving non-homogeneous linear system of differential equations}

\label{subsection:solve_ode}

\begin{figure}
  \centering
  \includegraphics[width=1.0\linewidth]{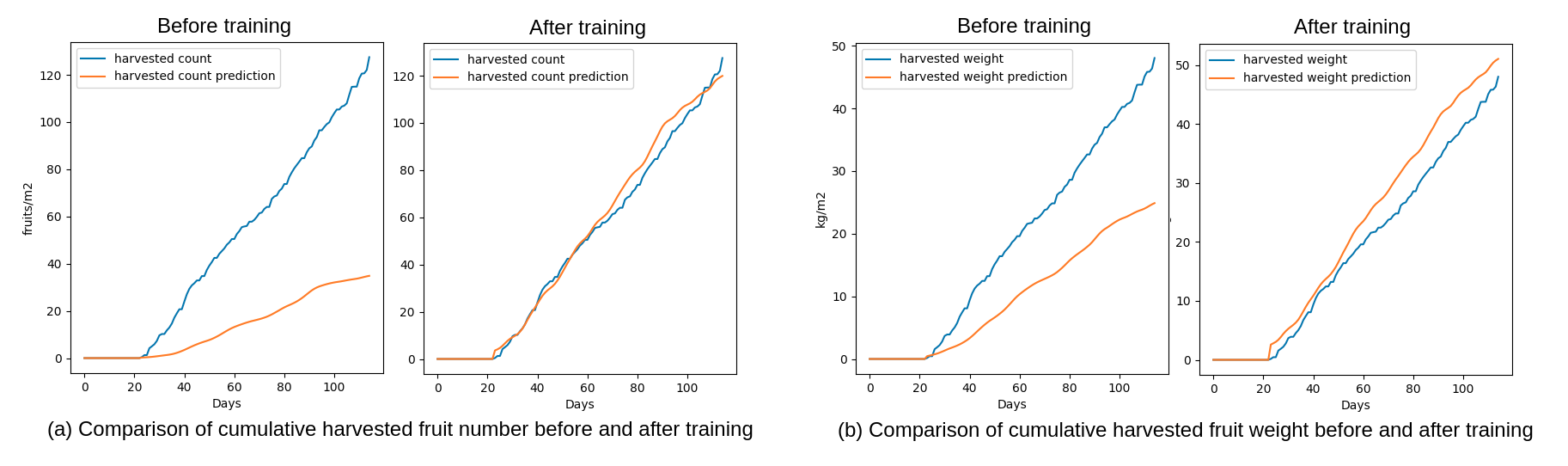}
  \caption{Crop prediction performance before and after training on the Croperators dataset. Blue: Ground truth. Orange: Our physics-based simulator prediction. Left: Cumulative harvested fruit number. Right: Cumulative harvested fruit weight.}
  \label{fig:harvest_plot}
\end{figure}

Suppose that $\bm{x}$ follows the dynamics $\dot{\bm{x}}=A\bm{x}+b$ where $A$ and $b$ are constants.
Denote $\bm{x}(t)$ as the state vector at time $t$.
We want to find $\bm{x}(t+h)$ given $\bm{x}(t)$, $h$ is the time step. 
The non-homogeneous linear system of differential equations has a closed form solution:
\begin{equation}
    \label{eq:ode_solve}
    \bm{x}(t+h) = e^{hA}\bm{x}(t) + \left(\int_{0}^{h}e^{(h-\tau)A}d\tau\right)b = e^{hA}\bm{x}(t) + \left(\int_{0}^{h}e^{\tau A}d\tau\right) b
\end{equation}

To obtain the next state $\bm{x}_{k+1}$ from $\bm{x}_{k}$, we substitute $\bm{x}_{k}$ for $\bm{x}(t)$, $\bm{x}_{k+1}$ for $\bm{x}(t+h)$ and set $h=\Delta t$.
In our simulation model, $A$ and $b$ are functions of the state $\bm{x}$ and the control $\bm{u}$, which change over time.
When $h$ is small enough, $A$ and $b$ can be treated as constants and we can use (\ref{eq:ode_solve}) for our simulation model.
The first matrix exponential term in (\ref{eq:ode_solve}) can be approximated by Taylor expansion:
\begin{equation}
    \label{eq:taylor_expansion}
    e^{X} = \sum_{j=0}^\infty \frac{(X)^j}{j!}=\sum_{j=0}^\infty C_j \text{ where } C_j =\frac{1}{j}\times C_{j-1}\times X \text{ for } j > 0 \text{ and } C_0=1
\end{equation} 
For computational purposes, we follow this recursive formulation until the change is less than $10^{-6}$ with a maximum of ten Taylor expansion terms. 

\par


The integral
$
    Q =\int_{0}^{h}e^{\tau A}d\tau = h\int_{0}^{1}e^{\tau(hA)}d\tau
$
in the second term of (\ref{eq:ode_solve}) can be computed exactly in a numerically stable fashion as follow.
Let
$   
Z = \begin{bmatrix}
    \begin{aligned}
    & hA  & I \\
    & 0  & 0
    \end{aligned}
    \end{bmatrix}
$
where $I$ is the identity matrix and $0$ are zero matrices so that $Z$'s dimension is twice that of $A$. The matrix exponential of $Z$ can be proved, by Taylor expansion, to be
$
    e^Z= \begin{bmatrix}
\begin{aligned}
& e^{hA}  & \int_0^1 e^{\tau(hA)}d\tau \\
& 0  & I
\end{aligned}
\end{bmatrix}
$.
Hence, we can obtain $Q/h$ by computing the matrix exponential of $Z$ using the same procedure in (\ref{eq:taylor_expansion}) and slice it.

\section{Experiment}
\label{section:experiment}

\subsection{Dataset}
\label{subsection:dataset}
We used the public dataset \footnote{Available at this  \href{https://github.com/CEAOD/Data/tree/master/GH_Cucumber/AutonomousGreenhouseChallenge2018}{Github repository}.} from the Autonomous Greenhouse Challenge 2018 \cite{Hemming2019} to train our simulator's parameters and test the results. The original dataset contains data on greenhouse climate, irrigation, outdoor weather, control setpoints, crop management, and resource consumption. The dataset has a time step of $\Delta t$ equal to five minutes. The collected data were from 4-month Hi-Power cucumber crops from six compartments of five participants (AiCU, Croperators, Deep Greens, iGrow, Sonoma) and a reference team, resulting in tens of thousands of data points total for each team. In this work, we trained on the AiCU dataset and validated on the Croperators dataset.

\subsection{Training details}
\label{subsection:training_details}
A sketch of the general training procedure is available in Algorithm \ref{algo:general_training}.
We considered the tasks of predicting the following quantities in the state vector: \textbf{(1)} Climate states: air temperature, air vapor pressure, CO2 in the air. \textbf{(2)} Crop states: cumulative harvested fruit weight and cumulative harvested fruit number everyday.
Denote $\bm{x}_{k}^{Temp}$, $\bm{x}_{k}^{VP}$, $\bm{x}_{k}^{CO2}$, $\bm{x}_{k}^{HW}$, $\bm{x}_{k}^{HC}$ as the aforementioned quantities of a state $\bm{x}_{k}$ in that order.
Furthermore, denote $\hat{\cdot}$ as the prediction of a state vector by the simulator. 
We consider the multiple-step prediction problem where the formula $\bm{x}_{k+1}=\mathcal{S}_{\theta}(\bm{x}_{k},\bm{u}_{k})$ is applied repeatedly to obtain the multistep prediction $\hat{\bm{x}}_{i+1}, \hat{\bm{x}}_{i+2},..., \hat{\bm{x}}_{i+T}$ from the starting state $\bm{x}_{i}$ and the known control vectors $\bm{u}_{i}, \bm{u}_{i+1}, ... \bm{u}_{i+T-1}$.
In practice, however, we don't have $\bm{x}_{i}$ because many states are unobservable. 
To solve this issue, we replace the starting $\bm{x}_{i}$ with $\tilde{\bm{x}}_{i}$ which is a combination of available data of the true states $\bm{x}_{i}$ and some approximation of the missing states computed by a reference model $\mathcal{S}_{ref}$.
Exactly how to compute $\tilde{\bm{x}}_{i}$ is described in section \ref{subsection:computing_approx_states}.
\par
The simulation model was trained using gradient descent with backpropagation \cite{kelley1960gradient}. 
The Adam optimizer \cite{kingma2014adam} with a learning rate of $10^{-1}$ and betas equal to $(0.9, 0.999)$ was used.
The absolute values of the gradients were clipped to a maximum of 1.0.
We separated the training processes of the climate states and crop states losses due to their differences in sampling scheme to compute the losses. 
For the climate states loss, we randomly sampled the starting state $\tilde{\bm{x}}_{i}$, used the simulator to compute the next $T$ states prediction, then computed the mean squared error loss for the prediction trajectory:
\begin{equation}
    \label{eq:loss_climate}
    \mathcal{L}_1(\theta,i)=\frac{1}{T} \sum_{k=i+1}^{i+T}\left[\omega^{Temp} (\hat{\bm{x}}_{k}^{Temp} - \bm{x}_{k}^{Temp})^2 + \omega^{VP} (\hat{\bm{x}}_{k}^{VP} - \bm{x}_{k}^{VP})^2 + \omega^{CO2} (\hat{\bm{x}}_{k}^{CO2} - \bm{x}_{k}^{CO2})^2 \right]
\end{equation}
where $\omega^{Temp}=10^{-1},\omega^{VP}=5*10^{-6},\omega^{CO2}=2*10^{-5}$.
We sampled multiple starting states at the same time for minibatch training of size 32. 
The prediction trajectory length $T$ was 24 (2 hours).
\par
The crop states training procedure is similar to that of the climate states, with a different loss:
\begin{equation}
    \label{eq:loss_crop}
    \mathcal{L}_2(\theta,i)=\sum_{k}\left[\omega^{HW} (\hat{\bm{x}}_{k}^{HW} - \bm{x}_{k}^{HW})^2 + \omega^{HC} (\hat{\bm{x}}_{k}^{HC} - \bm{x}_{k}^{HC})^2 \right]
\end{equation}
where $\omega^{HW}=\omega^{HC}=1$.
Because the harvested fruit weight and harvested fruit numbers data are only available daily, the index $i$ of the starting state was chosen to be start-of-day, the prediction length $T$ was chosen so that the last timestep is end-of-day and the indices $k$ were all the end-of-day timesteps in the prediction horizon. 
We used $T=2016$ (equal 7 days). The mini-batch size was 32.

\SetKwInput{KwInput}{Input}                
\SetKwInput{KwOutput}{Output}              

\begin{algorithm}[!ht]
\caption{General Training Procedure}
\label{algo:general_training}
\DontPrintSemicolon
  
  Initialize parameter vector $\theta$ \tcp*{Randomly in physically feasible ranges}
  
  \For{E iterations}{
      \tcp{E=10 in the experiment}
      
      Set $\mathcal{S}_{ref}$ to be the latest version of $S_{\theta}$
    
      Simulate the whole growing season with the reference model $\mathcal{S}_{ref}$ to compute all $\tilde{\bm{x}}_{i}$
      
      \For {$L_{climate}$ iterations}
      {
        \tcp{$L_{climate}$=50 in the experiment}
        
        Sample a random mini-batch of initial states $\tilde{\bm{x}}_{i}$ and control vectors $\bm{u}$
        
        Simulate using $\mathcal{S}_{ref}, \tilde{\bm{x}}_{i}, \bm{u}$,  to compute the prediction trajectories
        
        Compute the loss function in \ref{eq:loss_climate}
        
        Update $\theta$ using backpropagation
      }
      
      Set $\mathcal{S}_{ref}$ to be the latest version of $S_{\theta}$
    
      Simulate the whole growing season with the reference model $\mathcal{S}_{ref}$ to compute $\tilde{\bm{x}}_{i}$
      
      \For {$L_{crop}$ iterations}
      {
        \tcp{$L_{crop}$=25 in the experiment}
        
        Sample a mini-batch of initial states $\tilde{\bm{x}}_{i}$ and control vectors $\bm{u}$ using the sampling scheme associated with the loss \ref{eq:loss_crop}
        
        Simulate using $\mathcal{S}_{ref}, \tilde{\bm{x}}_{i}, \bm{u}$,  to compute the prediction trajectories
        
        Compute the loss function in \ref{eq:loss_crop}
        
        Update $\theta$ using backpropagation
      }

  }
\end{algorithm}

\subsection{Compute the state approximation $\tilde{\bm{x}}_{i}$}
\label{subsection:computing_approx_states}
At the start of the season, the missing states are initialized using other available state that is closest to it. 
For example, temperature of the attic air will be initialized using values of the general air temperature.
We then use the reference greenhouse simulation model $\mathcal{S}_{ref}$ to simulate the whole growing season using the known control vectors $\bm{u}$.
This gives predictions of the missing states. 
The predictions of the missing states are then combined with the states available in the data to make $\tilde{\bm{x}}_{i}$.
The intuition is that over a long time, the missing states should depend mostly on the control and not on their initialization.
Because this computation is very expensive, we only do this occasionally during the training process.
$\mathcal{S}_{ref}$ is chosen to be the physics-based greenhouse simulation model but with an old parameters vector and is updated periodically to reflect the updated parameters.

\subsection{Result}

We show some preliminary evaluation of our model in this paper. 
Figure \ref{fig:climate_loss_plot} shows the component of the loss (\ref{eq:loss_climate}) on the Croperator validation dataset. 
All the individual climate states losses and the total loss trended down over time. 
Figure \ref{fig:harvest_plot} shows the comparison between the ground truth and the prediction by our simulation model before and after training on the Croperator dataset.
The loss terms in (\ref{eq:loss_crop}) decreased from over 2500 to 400 for harvested fruit numbers and from 155 to just over 20 for harvested fruit weight.
Furthermore, for the prediction performance on the last day of the season (which matters the most for economic purpose), the relative errors decreased from 48\% to 6\% for predicting harvested fruit weight and from 73\% to 6\% for harvested fruit numbers.
Hence, we can conclude that the training significantly improved the simulation model to have good performance. 

\par

\section{Conclusion}
\label{section:conclusion}
We presented an interpretable and differentiable physics-based greenhouse simulation model and proposed an efficient inference and training procedure for it. 
The work has the potential to improve optimal control for greenhouses. 
Future directions are to experiment with more crop types and to exploit the fact that the crop dynamics changes slower than the climate dynamics to improve speed. 
\par
\textbf{Limitations of the work:} The work is still in progress and lacks rigorous experiments and further experiments on more crop types, such as tomato and lettuce.
\textbf{Potential negative societal impacts:} Automation of greenhouse simulation and control may take away jobs from human experts.
However, we argue that this will free up human labor capital.
Moreover, experts are still needed to better tune the simulator models and the optimal controllers.



\section*{Broader Impacts}
Our work has the potential to facilitate improvement and allow faster deployment of optimal control algorithms for greenhouses. It is also a proof of concept that a physics-based greenhouse simulation model can be trained end-to-end to achieve good performance. Better automation of greenhouse control can help address the food security issue that many countries currently face. The procedure to solve the differential system of differential equations and the training procedure in this work can also be applied to similar systems. 

\bibliographystyle{plain}
\bibliography{references}

\section*{Checklist}

\begin{enumerate}

\item For all authors...
\begin{enumerate}
  \item Do the main claims made in the abstract and introduction accurately reflect the paper's contributions and scope?
    \answerYes{}
  \item Did you describe the limitations of your work?
    \answerYes{} See section \ref{section:conclusion}
  \item Did you discuss any potential negative societal impacts of your work?
    \answerYes{} See section \ref{section:conclusion}
  \item Have you read the ethics review guidelines and ensured that your paper conforms to them?
    \answerYes{}
\end{enumerate}

\item If you are including theoretical results...
\begin{enumerate}
  \item Did you state the full set of assumptions of all theoretical results?
    \answerNA{}
        \item Did you include complete proofs of all theoretical results?
    \answerNA{}
\end{enumerate}

\item If you ran experiments...
\begin{enumerate}
  \item Did you include the code, data, and instructions needed to reproduce the main experimental results (either in the supplemental material or as a URL)?
    \answerNo{} The code is proprietary.
  \item Did you specify all the training details (e.g., data splits, hyperparameters, how they were chosen)?
    \answerYes{}
        \item Did you report error bars (e.g., with respect to the random seed after running experiments multiple times)?
    \answerNo{}
        \item Did you include the total amount of compute and the type of resources used (e.g., type of GPUs, internal cluster, or cloud provider)?
    \answerNA{} No GPUs was used. The model is very small and training can be done on CPU in a couple of hours.
\end{enumerate}

\item If you are using existing assets (e.g., code, data, models) or curating/releasing new assets...
\begin{enumerate}
  \item If your work uses existing assets, did you cite the creators?
    \answerYes{}
  \item Did you mention the license of the assets?
    \answerYes{}
  \item Did you include any new assets either in the supplemental material or as a URL?
    \answerNo{}
  \item Did you discuss whether and how consent was obtained from people whose data you're using/curating?
    \answerNA{}
  \item Did you discuss whether the data you are using/curating contains personally identifiable information or offensive content?
    \answerNA{}
\end{enumerate}

\item If you used crowdsourcing or conducted research with human subjects...
\begin{enumerate}
  \item Did you include the full text of instructions given to participants and screenshots, if applicable?
    \answerNA{}
  \item Did you describe any potential participant risks, with links to Institutional Review Board (IRB) approvals, if applicable?
    \answerNA{}
  \item Did you include the estimated hourly wage paid to participants and the total amount spent on participant compensation?
    \answerNA{}
\end{enumerate}

\end{enumerate}




\end{document}